\documentclass[graybox]{svmult}

\usepackage{mathptmx}       % selects Times Roman as basic font
\usepackage{helvet}         % selects Helvetica as sans-serif font
\usepackage{courier}        % selects Courier as typewriter font
\usepackage{type1cm}        % activate if the above 3 fonts are
\usepackage{makeidx}         % allows index generation
\usepackage{graphicx}        % standard LaTeX graphics tool
\usepackage{multicol}        % used for the two-column index
\usepackage[bottom]{footmisc}% places footnotes at page bottom
\usepackage{gensymb}
\usepackage{amssymb}
\usepackage[caption=false,font=footnotesize]{subfig}
\usepackage{amsmath}
\usepackage{subfig}

\makeindex             % used for the subject index
\begin{document}

%\title*{Risk-aware Planning for a Collaborative Tethered Aerial Visual Assistant for Robot Operations in Unstructured or Confined Environments}
\title*{Autonomous Visual Assistance for Robot Operations Using a Tethered UAV}
\author{Xuesu Xiao, Jan Dufek, and Robin R. Murphy}
\institute{Xuesu Xiao \and Jan Dufek \and Robin R. Murphy \\ \email{\{xiaoxuesu\}, \{dufek\}, \{robin.r.murphy\}@tamu.edu}
\at Department of Computer Science and Engineering, Texas A\&M University}

\maketitle

\abstract*{
% objectives with no jargon
This paper develops an autonomous tethered aerial visual assistant for robot operations in unstructured or confined environments. 
% how it is done today
Robotic tele-operation in remote environments is difficult due to lack of sufficient situational awareness, mostly caused by the stationary and limited field-of-view and lack of depth perception from the robot's onboard camera. The emerging state of the practice is to use two robots, a primary and a secondary that acts as a visual assistant to overcome the perceptual limitations of the onboard sensors by providing an external viewpoint. However, problems exist when using a tele-operated visual assistant: extra manpower, manually chosen suboptimal viewpoint, and extra teamwork demand between primary and secondary operators. 
% what is new
In this work, we use an autonomous tethered aerial visual assistant to replace the secondary robot and operator, reducing human robot ratio from 2:2 to 1:2. This visual assistant is able to autonomously navigate through unstructured or confined spaces in a risk-aware manner, while continuously maintaining good viewpoint quality to increase the primary operator's situational awareness. 
% Who cares
With the proposed co-robots team, tele-operation missions in nuclear operations, bomb squad, disaster robots, and other domains with novel tasks or highly occluded environments could benefit from reduced manpower and teamwork demand, along with improved visual assistance quality based on trustworthy risk-aware motion in cluttered environments. }

\abstract{
% objectives with no jargon
This paper develops an autonomous tethered aerial visual assistant for robot operations in unstructured or confined environments. 
% how it is done today
Robotic tele-operation in remote environments is difficult due to lack of sufficient situational awareness, mostly caused by the stationary and limited field-of-view and lack of depth perception from the robot's onboard camera. The emerging state of the practice is to use two robots, a primary and a secondary that acts as a visual assistant to overcome the perceptual limitations of the onboard sensors by providing an external viewpoint. However, problems exist when using a tele-operated visual assistant: extra manpower, manually chosen suboptimal viewpoint, and extra teamwork demand between primary and secondary operators. 
% what is new
In this work, we use an autonomous tethered aerial visual assistant to replace the secondary robot and operator, reducing human robot ratio from 2:2 to 1:2. This visual assistant is able to autonomously navigate through unstructured or confined spaces in a risk-aware manner, while continuously maintaining good viewpoint quality to increase the primary operator's situational awareness. 
% Who cares
With the proposed co-robots team, tele-operation missions in nuclear operations, bomb squad, disaster robots, and other domains with novel tasks or highly occluded environments could benefit from reduced manpower and teamwork demand, along with improved visual assistance quality based on trustworthy risk-aware motion in cluttered environments. }

\section{Introduction}
\label{sec::introduction}
Tele-operated robots are still widely used in DDD (Dangerous, Dirty, and Dull) environments where human presence is extremely difficult or impossible, due to those environments' mission-critical task execution and current technological limitations. Projecting human presence to remote environments is still an effective approach to leverage current technologies and actual field demand. However, one major challenge of tele-operation is the insufficient situational awareness of the remote field, caused by the onboard sensing limitations, such as relatively stationary and limited field of view and lack of depth perception from the robot's onboard camera. Therefore, the emerging state of the practice for nuclear operations, bomb squad, disaster robots, and other domains with novel tasks or highly occluded environments is to use two robots, a primary and a secondary that acts as a visual assistant to overcome the perceptual limitations of the sensors by providing an external viewpoint.

However, the usage of tele-operated visual assistants also causes problems: it requires an extra human operator, or even operating crew, to tele-operate the secondary visual assistant. Human operators also tend to choose suboptimal viewpoints based on experience only. Most importantly, communication between the two operators of the primary and secondary robots requires extra teamwork demand, in addition to the task and perceptual demands of the tele-operation. In this research, an autonomous tethered aerial visual assistant is developed to replace the secondary robot and its operator, reducing the human robot ratio from 2:2 to 1:2. The co-robots team will then consist of one tele-operated primary ground robot, one autonomous aerial visual assistant, and one human operator, whose situational awareness is maintained by the visual feedback streamed from a series of optimal viewpoints for the particular tele-operation task. 

Unmanned Ground Vehicles (UGVs) are stable, reliable, durable, and can thus represent humans to actuate upon the real world, while Unmanned Aerial Vehicles (UAVs) have superior mobility and workspace coverage and therefore are capable of providing enhanced situational awareness \cite{murphy2016two}. Researchers have looked into utilizing the advantages and avoiding the disadvantages by teaming up the two types of robots \cite{chaimowicz2005deploying, cheung2008uav}. A more relevant area was to use a UAV to augment the UGV's perception or assist UGV's task execution, such as ``an eye in the sky'' for UGV localization \cite{chaimowicz2004experiments}, providing stationary third person view for construction machine \cite{kiribayashi2018design}, improving navigation in case of GPS loss \cite{frietsch2008teaming}, UGV control with UAV's visual feedback \cite{xiao2017uav, xiao2015locomotive} using differential flatness \cite{rao2003visual}. However, instead of prior works' flight path execution in wide open space or hovering at a stationary and elevated viewpoint, our aerial visual assistant needs to navigate through unstructured or confined spaces in order to provide visual assistance to the UGV operator from a series of good viewpoints. It is able to reason about the motion execution risk in complex environments and plan a path that provides good visual assistance. In particular, this work uses a tethered UAV, with the purpose of matching its battery duration with UGV's and as a failsafe in case of malfunction in mission-critical tasks. Our tethered visual assistant utilizes the advantages and mitigates the disadvantages of the tether, in terms of tether-based indoor localization, motion primitives, and environment contact planning.

The remainder of this article is organized as follows: Sec. \ref{sec::team} presents the heterogeneous co-robots team. The high level visual assistance components are described in Sec. \ref{sec::components}, while low level tether-based motion implementation in Sec. \ref{sec::tethered_motion}. System demonstrations are provided in Sec. \ref{sec::system_demonstration}. Sec. \ref{sec::conclusions} concludes the paper.

\section{Co-Robots Team}
\label{sec::team}
\begin{figure}
\centering
\sidecaption[]
\includegraphics[width=0.6\columnwidth]{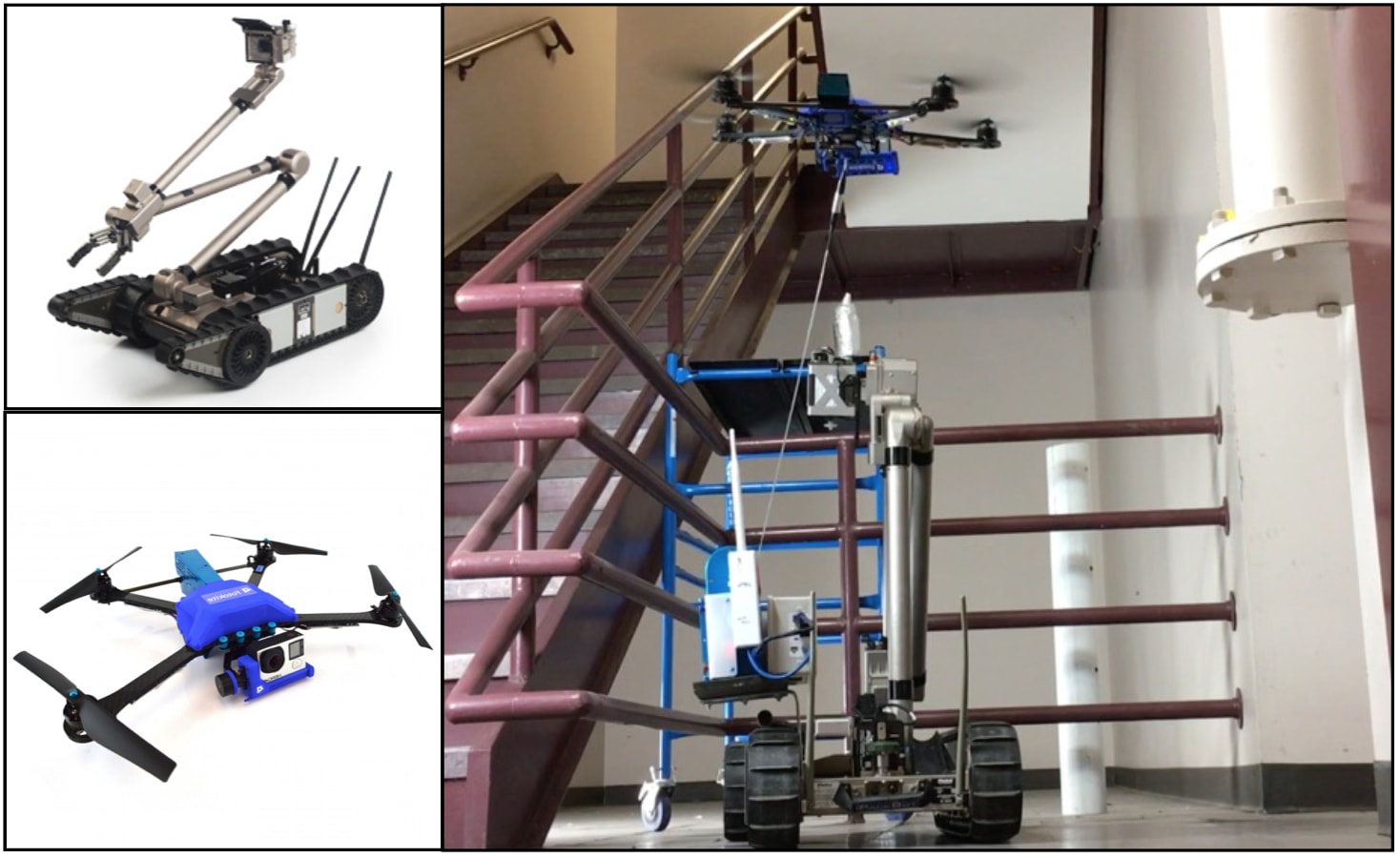}
\caption{The Co-Robots Team: tele-operated primary robot, Endeavor PackBot 510 (upper left), and autonomous tethered visual assistant, Fotokite Pro (lower left), picking up a sensor and dropping it into a radiation pipe in a confined staircase (right). }
\label{fig::team}
\end{figure}

This section presents the co-robots team: a tele-operated ground primary robot, an autonomous tethered aerial visual assistant, and a human operator of the primary robot under the visual assistance of the aerial vehicle (Fig. \ref{fig::team}). 

\subsection{Tele-Operated Ground Primary Robot}
In the co-robots team, the primary robot is a tele-operated Endeavor PackBot 510 (Fig. \ref{fig::team} upper left). PackBot has a chassis with two main differential treads that allow zero radius turn and maximum speed up to 9.3 km/h. Two articulated flippers with treads are used to climb over obstacles or stairs (up to 40\degree). PackBot's three-link manipulator locates on topic of the chassis, with an articulated gripper on the second link and an onboard camera on the third. The manipulator can lift 5kg at full extension and 20kg close-in. Motor encoders on the arm provide precise position of the articulated joints, including the gripper, the default visual assistance point of interest. Four onboard cameras provide first-person-views, but are all limited to the robot body. On the chassis, a Velodyne Puck LiDAR constantly scans the 3-D environments, providing the map for the co-robots team to navigate through. The map does not necessarily need to be global, with the unknown parts being assumed as obstacles. Four BB-2590 batteries provide up to 8 hrs run time. 

\subsection{Autonomous Aerial Visual Assistant}
A tethered UAV, Fotokite Pro, is used as the autonomous aerial visual assistant (Fig. \ref{fig::team} lower left). It could be deployed from a landing platform mounted on the ground robot's chassis. The UAV is equipped with an onboard camera with a 2-DoF gimbal (pitch and roll). The camera's yaw is controlled dependently by the vehicular yaw. The main purpose of the tether is to match the run time of the aerial visual assistant with the ground primary robot, since flight power could be transmitted via the tether. Additionally, tether serves as a fail-safe in mission-critical environments. The UAV's flight controller is based on the tether sensory feedback, including the tether length, azimuth and elevation angles. The six dimensional coverage of the workspace makes the UAV suitable for the visual assistance purpose. 

\subsection{Human Operator}
\begin{figure}
\centering
\subfloat[PackBot uPoint Controller Interface]{\includegraphics[width=0.43\columnwidth]{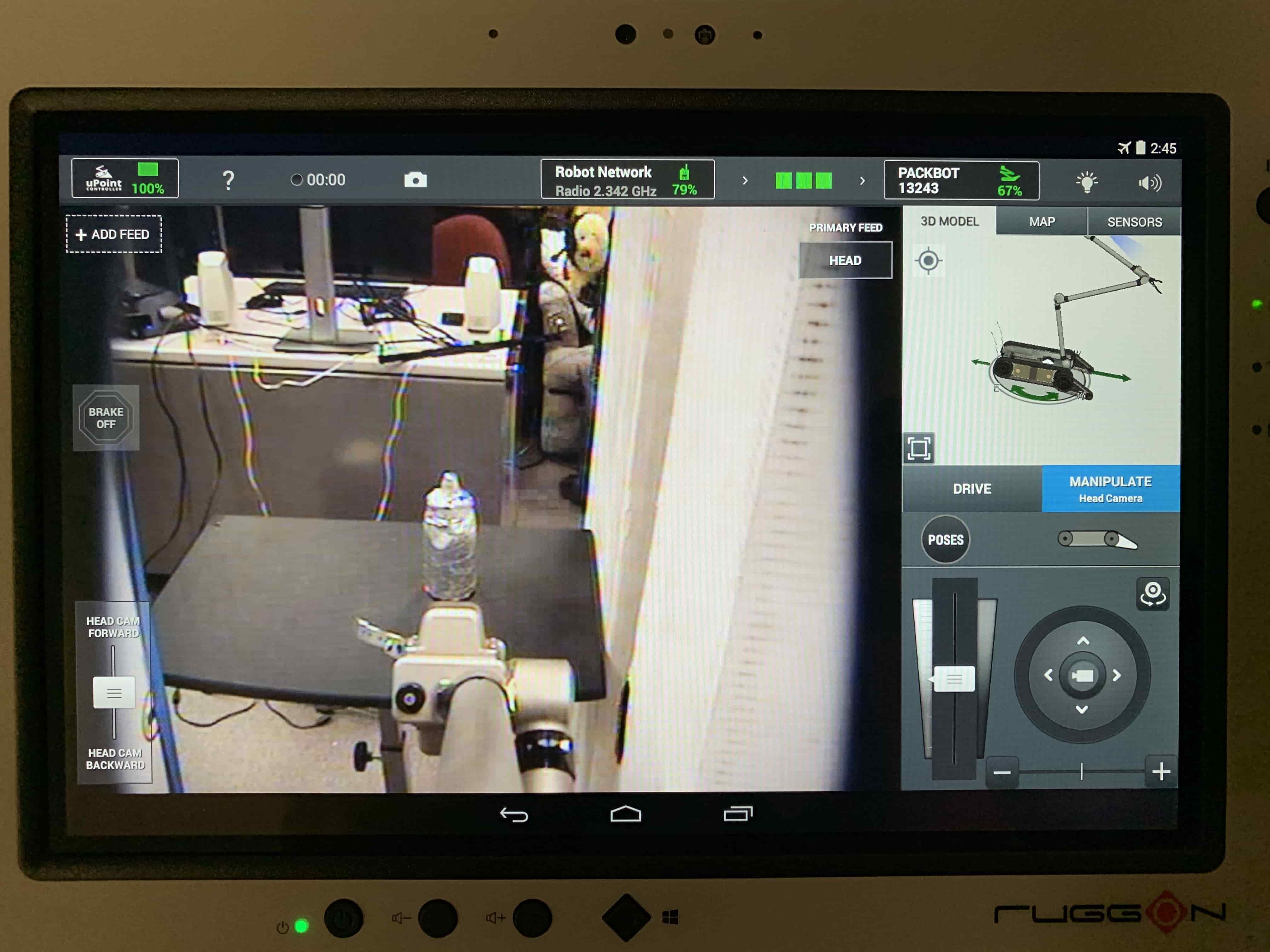}%
\label{fig::upoint}}
\subfloat[Visual Assistant Interface]{\includegraphics[width=0.57\columnwidth]{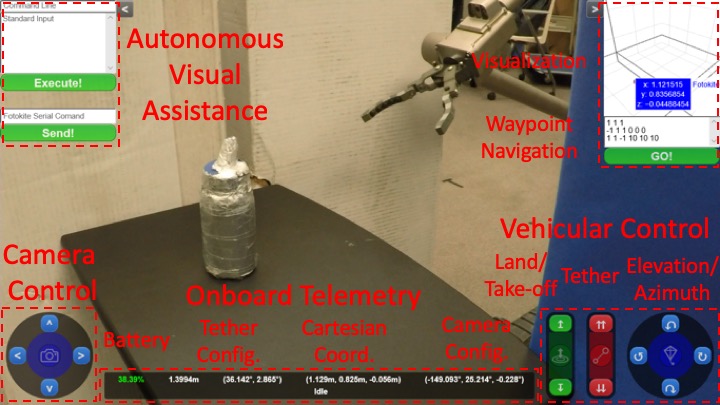}%
\label{fig::mickie}}
\caption{Interfaces with the Human Operator}
\label{fig::interfaces}
\end{figure}

The human operator tele-operates the primary ground robot with the visual assistance of the UAV. In addition to the default PackBot uPoint controller with onboard first-person-view, the visual feedback from the visual assistant's onboard camera is also available for enhanced situational awareness. For example, the visual assistant could move to a location perpendicular to the tele-operation action, providing extra depth perception to the operator. The visual assistant could be either manually controlled or automated. For the focus of this research, autonomous visual assistance, a 3-D map is provided by the primary robot's LiDAR, and a risk-aware path is planned using a pre-established viewpoint quality map (discussed in the following sections). The uPoint tele-operation and visual assistance interfaces are shown in Fig. \ref{fig::interfaces}

\section{Visual Assistance Components}
\label{sec::components}
This section introduces the key components of autonomous visual assistance, including a viewpoint quality map based on the cognitive science concept of affordances, an explicit path risk representation with a focus on unstructured or confined environments, and a planning framework to balance the trade-off between reward (viewpoint quality) and risk (motion execution). 

\subsection{Viewpoint Quality Reward}
\begin{figure}
\centering
\subfloat[Good Manipulability Viewpoint]{\includegraphics[width=0.45\columnwidth]{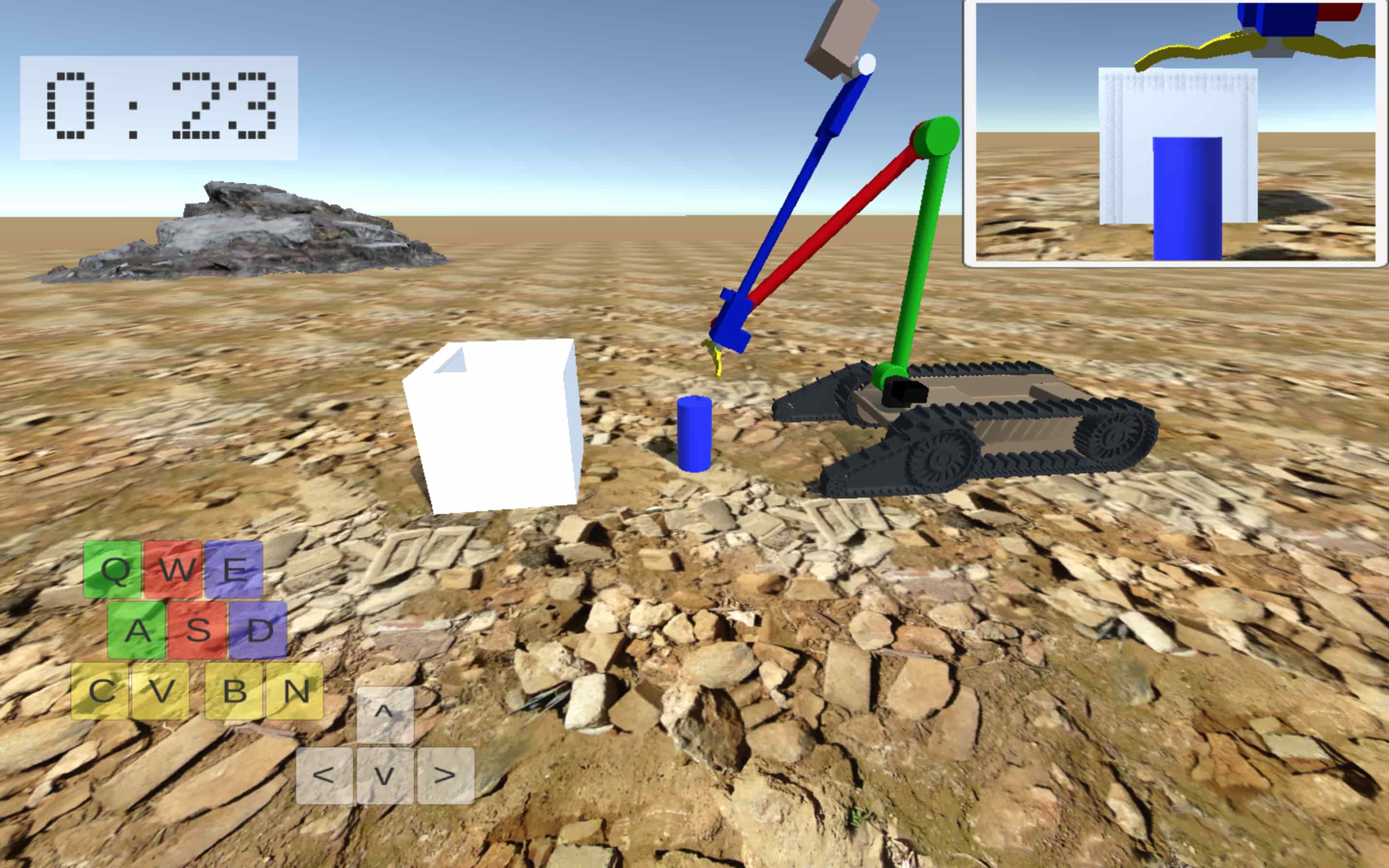}%
\label{fig::ManipulabilityGood}}
\hspace{10pt}
\subfloat[Bad Manipulability Viewpoint]{\includegraphics[width=0.45\columnwidth]{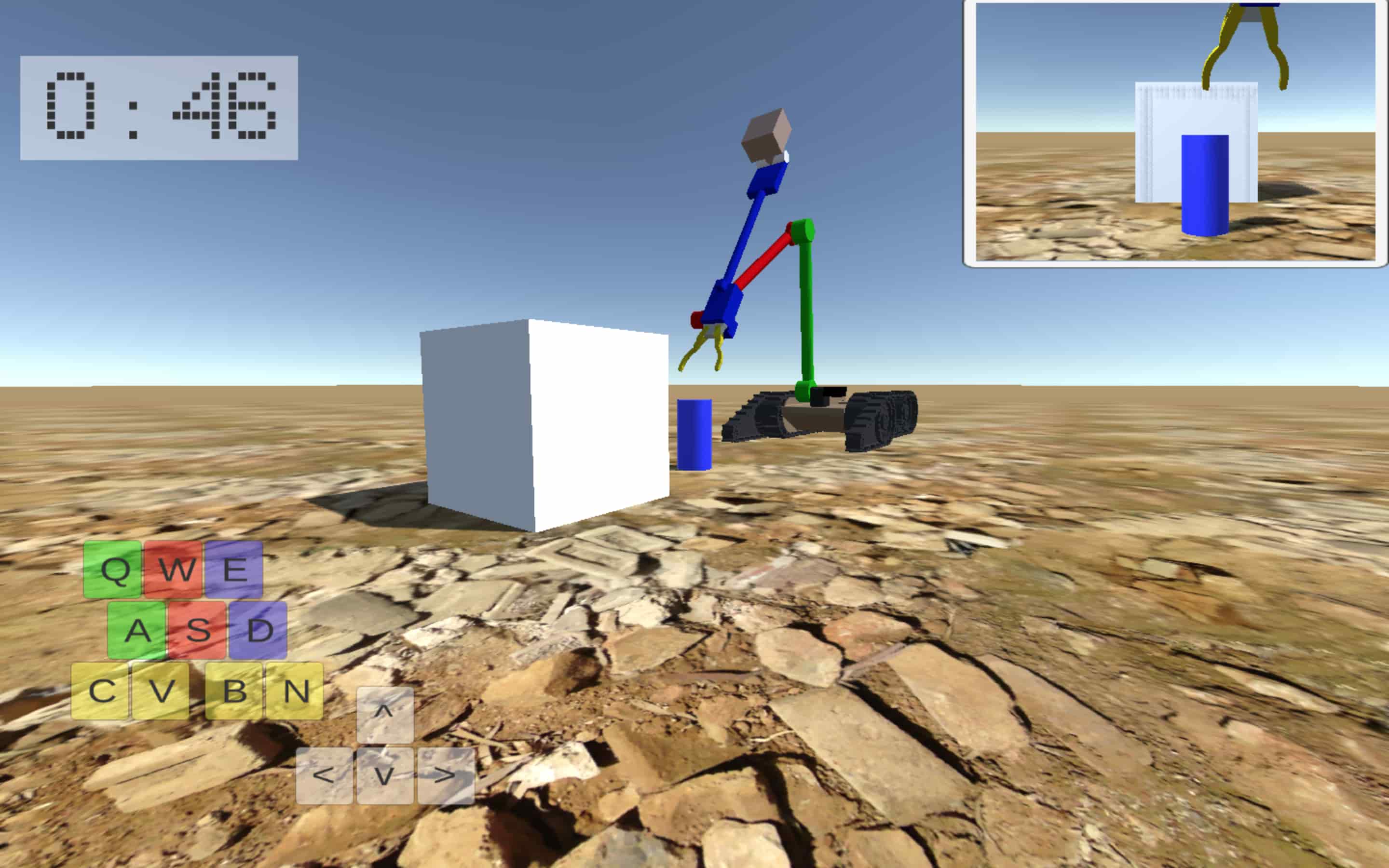}%
\label{fig::ManipulabilityBad}}\\
\subfloat[Good Passability Viewpoint]{\includegraphics[width=0.45\columnwidth]{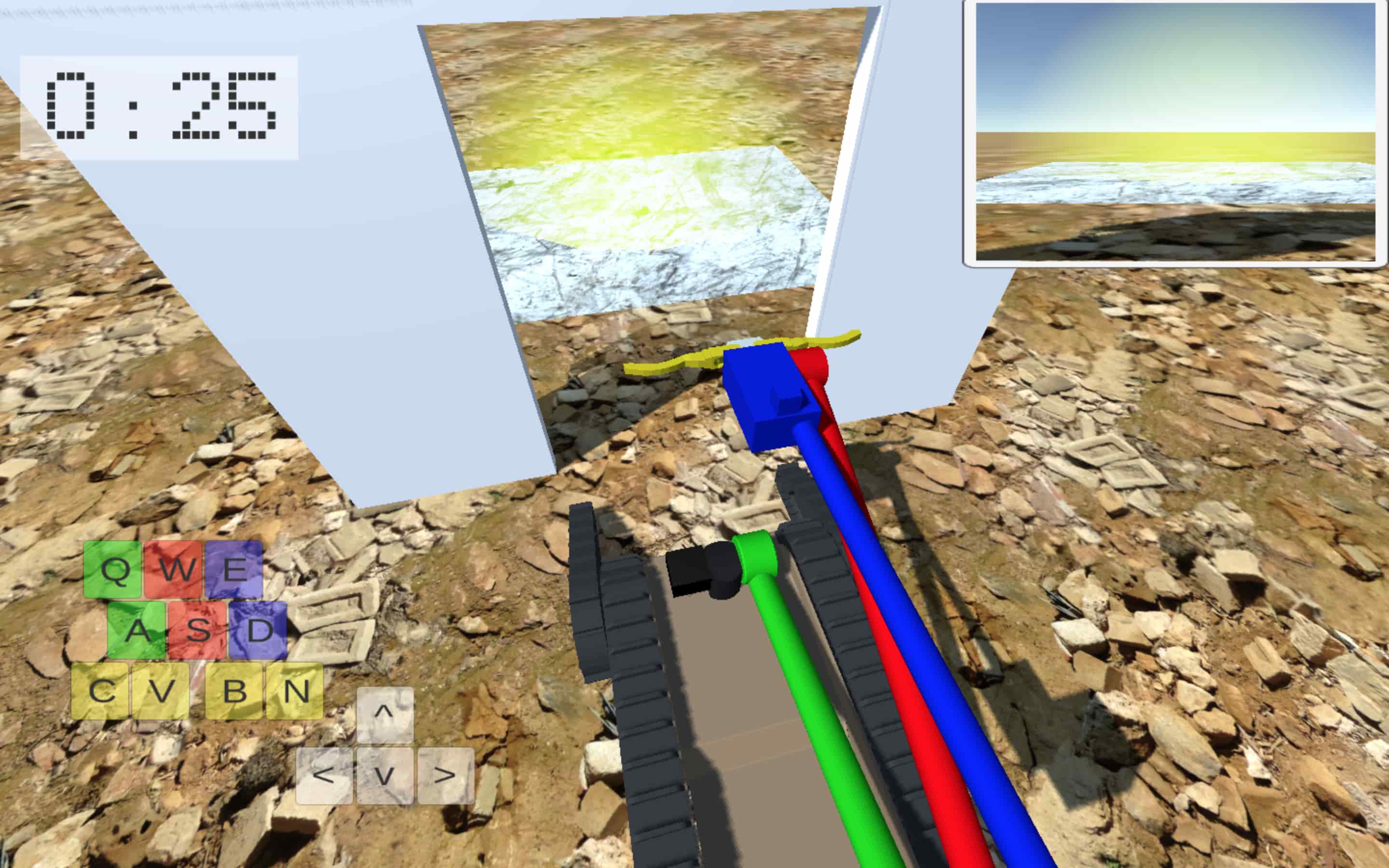}%
\label{fig::ManipulabilityGood}}
\hspace{10pt}
\subfloat[Bad Passability Viewpoint]{\includegraphics[width=0.45\columnwidth]{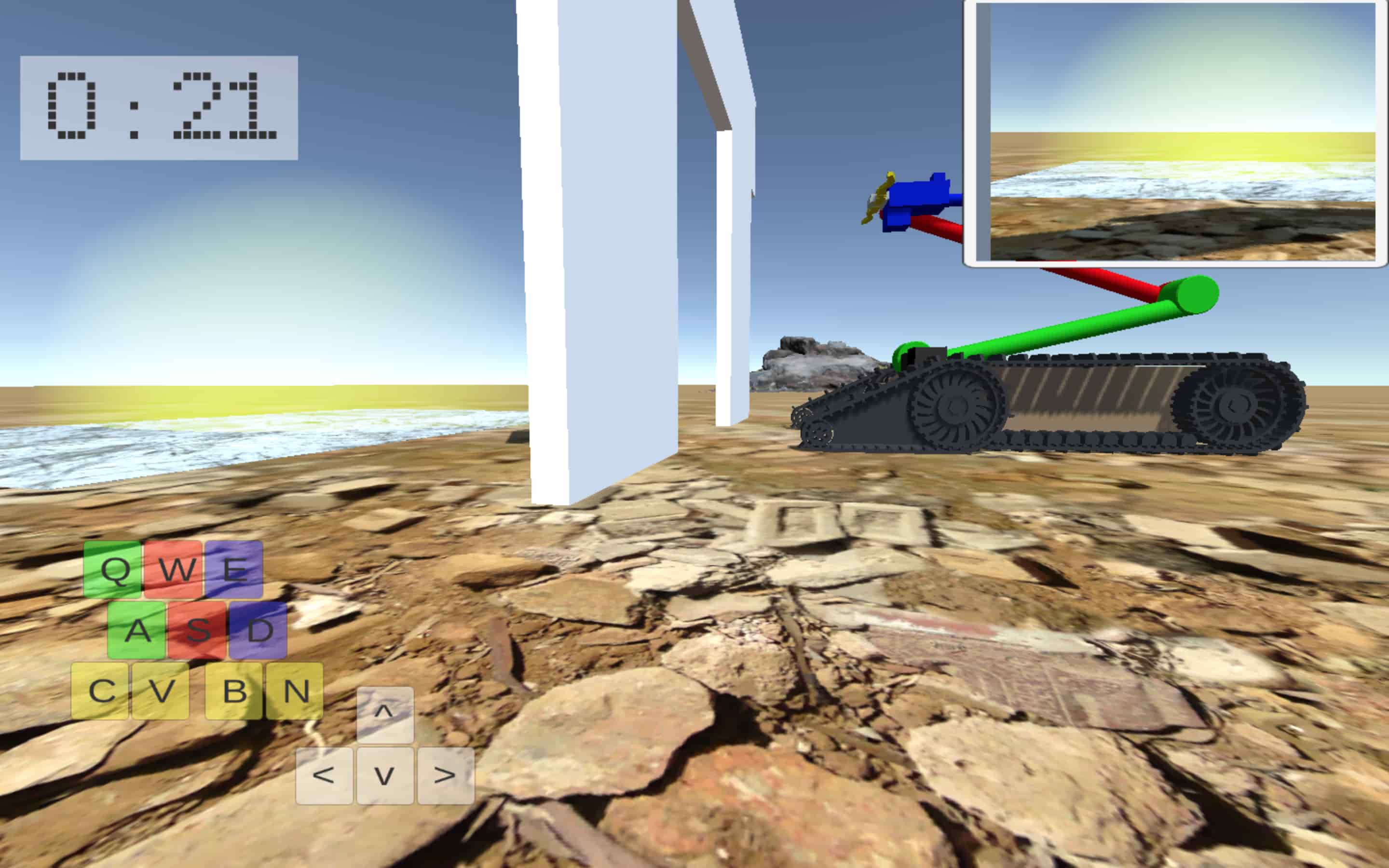}%
\label{fig::ManipulabilityBad}}\\
\subfloat[Good Reachability Viewpoint]{\includegraphics[width=0.45\columnwidth]{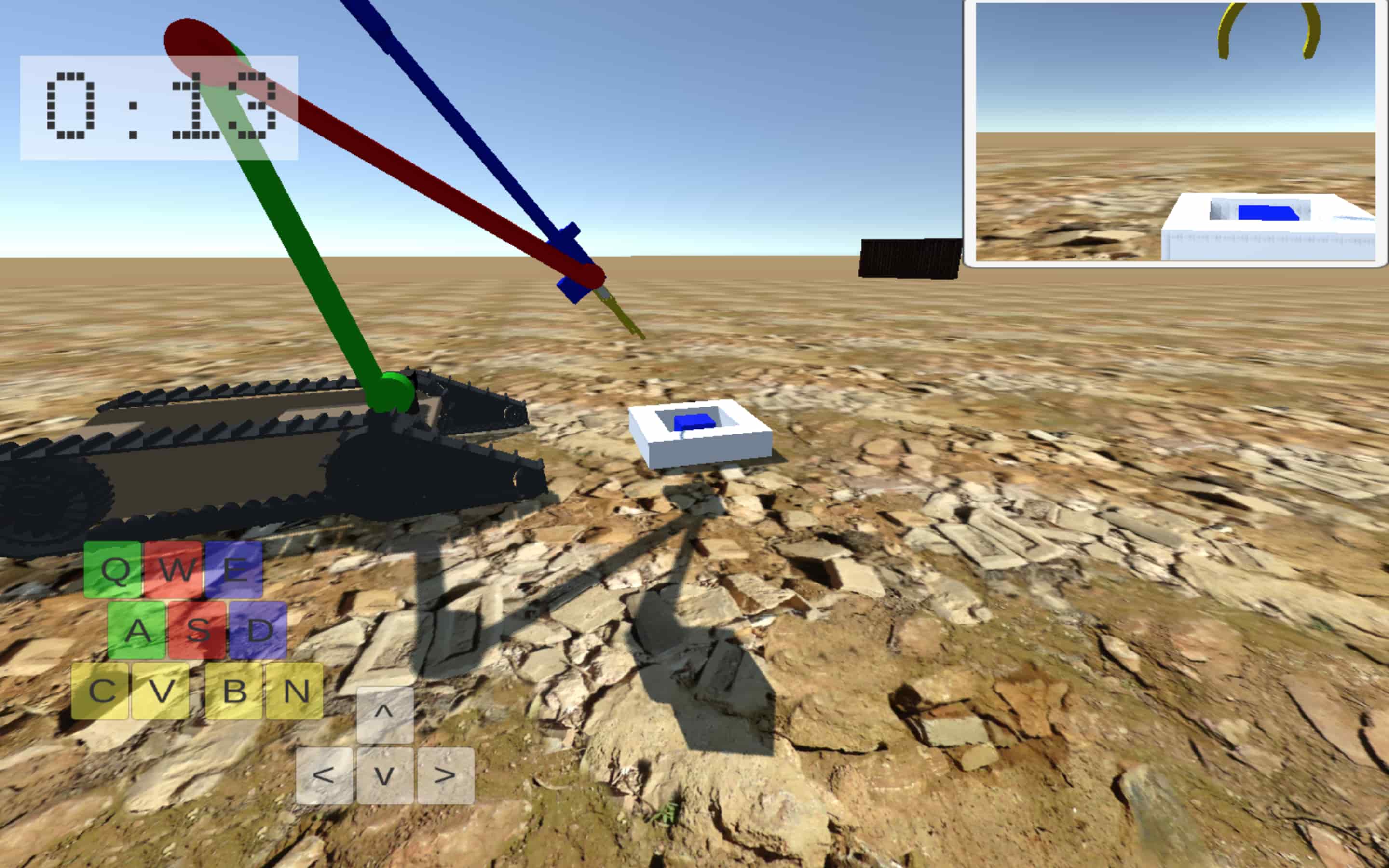}%
\label{fig::ManipulabilityGood}}
\hspace{10pt}
\subfloat[Bad Reachability Viewpoint]{\includegraphics[width=0.45\columnwidth]{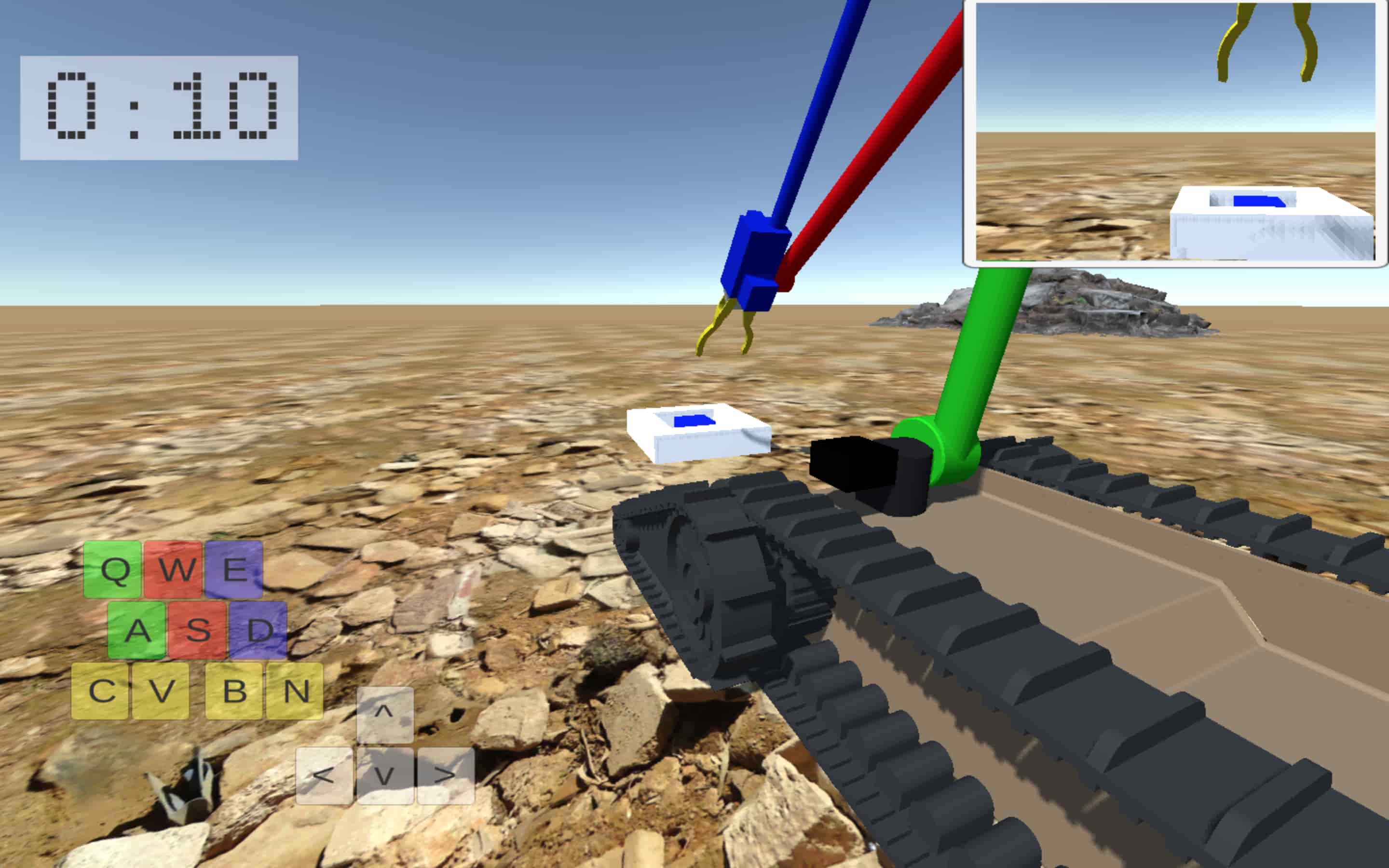}%
\label{fig::ManipulabilityBad}}\\
\subfloat[Good Traversability Viewpoint]{\includegraphics[width=0.45\columnwidth]{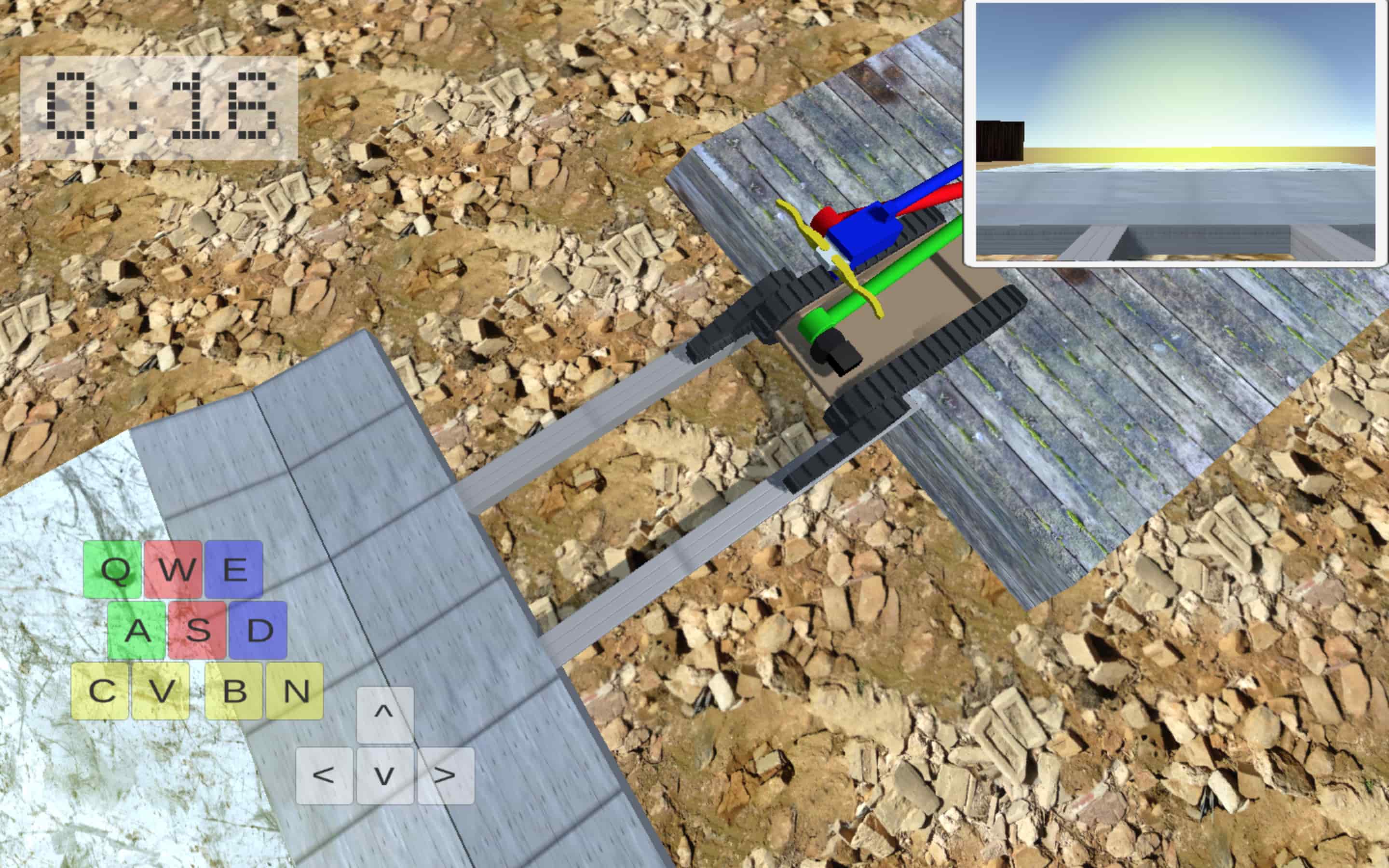}%
\label{fig::ManipulabilityGood}}
\hspace{10pt}
\subfloat[Bad Traversability Viewpoint]{\includegraphics[width=0.45\columnwidth]{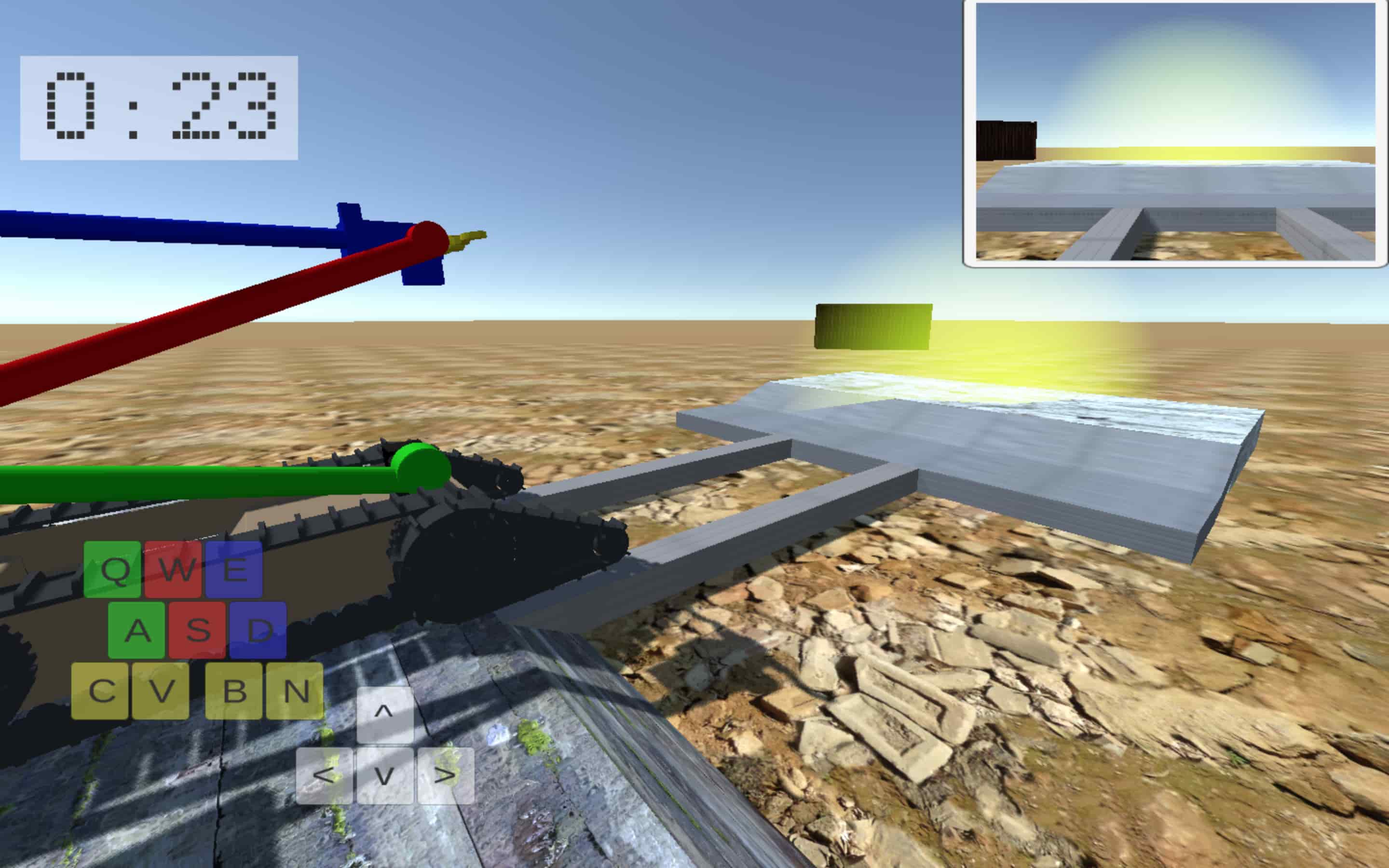}%
\label{fig::ManipulabilityBad}}\\
\caption{Ongoing Viewpoint Quality Study in Simulation with Professional PackBot Operators}
\label{fig::affordances}
\end{figure}

The cognitive science concept of affordances is used to determine viewpoint quality, where the potential for an action can be directly perceived without knowing intent or models, and thus is universal to all robots and tasks \cite{murphy2013apprehending}. For this work, four affordances are used: manipulability, passability, reachability, and traversability (Fig. \ref{fig::affordances}). In order to determine the viewpoint quality (reward) for each affordance, we use a computer simulation to collect performance data with 30 professional PackBot operators. A hemisphere centered at each affordance is created, with 30 viewpoints evenly distributed on it. The 30 viewpoints are divided into five groups: left, right, front, back, and above the affordance. For each affordance, every test subject is randomly given one viewpoint within each of the five groups and tries to finish the tele-operation task in an error-free and fast manner. The number of errors, such as colliding with the wall for passability or falling off the ledge for traversability, and the completion time are recorded. The average value of the product of error and time collected by all subjects is the metric to reflect the viewpoint quality. Given any point in the 3-D space, its viewpoint  reward is assigned to be the viewpoint quality of the closest point on the hemisphere. This study is still ongoing and its results will be reported in a separate paper.

\subsection{Explicit Risk Representation}
In contrast to the traditional state-dependent risk representation or probabilistic uncertainty modeling, this work uses an explicit risk representation as a function of entire path. The workspace $\mathbb{W}$ of the robot could be constructed by an occupancy grid from the LiDAR, excluding a set of obstacles $\mathbb{OB}= \{ob_i | i = 1, 2, ..., o\}$, where $o$ is the number of obstacles. Given a start location of the visual assistant $S$, a simple path $P$ could be defined as $P = \{s_0, s_1, ..., s_n\}$ where $s_i$ denotes the $i$th state on the path $P$ while $s_0 = S$, $ \forall 1\leq i\leq n, 1\leq j\leq o, s_i\cap ob_j = \emptyset$, and $\forall i\neq j, 1\leq i, j \leq n, s_i \neq s_j$. In a conventional state-dependent risk representation, risk at state $s_i$ is defined based on a function mapping from one state to a risk index $r: s_i \mapsto ri$ and the risk of executing a path $P$ is a simple summation of all individual states $risk(P) = \sum_{i=1}^{n} r(s_i)$. In the proposed path-dependent risk representation, however, risk at state $s_i$ cannot be simply evaluated by the state alone, but also the path leading to $s_i$, $P_i = \{s_0, s_1, ..., s_i\}$ and the risk at $s_i$ is computed through the mapping $R: (s_0, s_2, ..., s_i) \mapsto ri$. The path-level risk is relaxed from the simple summation to a more general representation $risk(P) = risk(s_0, s_1, ..., s_i)$. 

\begin{figure}
\centering
\includegraphics[width=1\columnwidth]{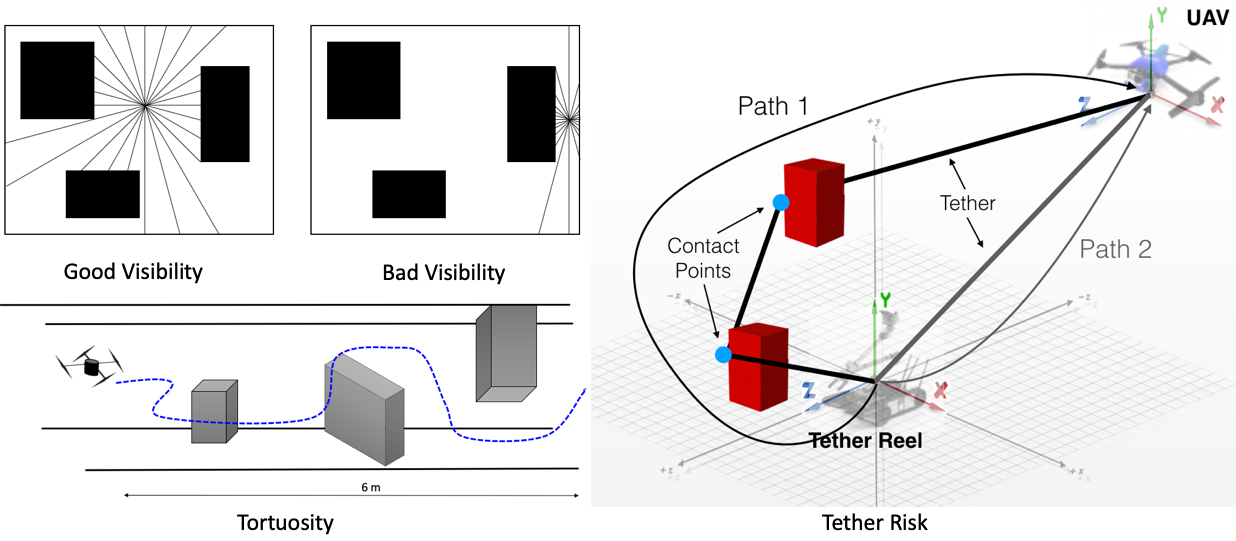}
\caption{Examples of Risk Elements: visibility (upper left), tortuosity (lower left), tether length, number of contacts, and azimuth (right)}
\label{fig::risk_elements}
\end{figure}

Our explicit path risk representation does not exclude the traditional state-dependent risk elements. Those risk functions $r: s_i \mapsto ri$ include risk elements caused by distance to closest obstacles $r_{di} = dis(s_i)$, visibility from isovists lines (Fig. \ref{fig::risk_elements} upper left) $r_{vi} = vis(s_i)$, altitude due to propeller ceiling or ground effect $r_{ai} = alt(s_i)$, and tether singularity $r_{si} = sig(s_i)$. Those risk elements are additive along the path. Path-dependent risk elements include action length, access element, tortuosity, number of tether contacts, tether length, and azimuth (Table \ref{tab::path_depedent_risk}). In order to define these risk elements, we further define the action between two consecutive states $s_{i-1}$ and $s_{i}$ to be $A_i$. So the whole sequence of actions to execute the entire path $P$ is defined as $\mathbb{A} = \{A_1, A_2, ..., A_n\}$. For action length, $\lVert A_i \rVert$ denotes the length of executing action $A_i$. For access elements, the function $AE$ evaluates the difficulty of entering from the void where $s_{i-1}$ is located to the void of $s_i$. Only positive difficulty is added into the risk index. Tortuosity characterizes the the number of ``turns'' necessary to reach the state. More generally speaking, this is the difference by some measurement between two consecutive actions $\lVert A_i - A_{i-1} \rVert$. Tether length is a function of entire path, e.g., taking path 1 and path 2 in Fig. \ref{fig::risk_elements} right will have completely different tether length. Number of contact points and azimuth angle are also different. Risk index should never decrease with the execution of a path, which is guaranteed by the norm and $max$ operations for the first three elements in Table \ref{tab::path_depedent_risk}. Thus they only need to be evaluated once for each path (unitary). For the last three, however, risk associated with each state may decrease, e.g., contact points may be relaxed \cite{xiao2018motion} and tether length may decrease. But this does not cancel the previous risk. Therefore those three elements need to be added for all states. Given a path $P$, its execution risk could be evaluated based on each risk element. Weighted sum or fuzzy logic could be used to combine all elements into one total risk index, quantifying the difficulty of executing that path. Detailed information of the explicit risk representation could be found in \cite{xiao2019explicit2}. 

%\begin{itemize}
%\item{Action Length: $R_{AC}(P) = \sum\limits_{i = 1}^{n} \lVert A_i \rVert$}
%
%\item{Access Element: $R_{AE}(P) = \sum\limits_{i = 2}^{n} max(AE(void(s_i), void(s_{i-1})), 0) $}
%
%\item{Tortuosity: $R_{T}(P) = \sum\limits_{i = 2}^{n} \lVert A_i - A_{i-1} \rVert$}
%
%\item{Tether Length: $R_{TL}(P)  = f(s_0, s_1, ..., s_n)$}
%
%\item{Number of Contacts: $R_{NC}(P)  = g(s_0, s_1, ..., s_n)$}
%
%\item{Azimuth: $R_{A}(P)  = h(s_0, s_1, ..., s_n)$}
%
%\end{itemize}

\begin{table}
\caption{Path-dependent Risk Elements}
\label{tab::path_depedent_risk} 
\begin{tabular}{p{0.3\columnwidth}p{0.6\columnwidth}p{0.1\columnwidth}}
\hline
Risk Element       & Risk Index & Property \\ \hline
Action Length      &      $R_{AL}(P) = \sum_{i = 1}^{n} \lVert A_i \rVert$      &    Unitary      \\ 
Access Element     &     $R_{AE}(P) = \sum_{i = 2}^{n}   max(AE(void(s_{i-1}), void(s_{i})), 0) $       &     Unitary     \\ 
Tortuosity         &      $R_{T}(P) = \sum_{i = 2}^{n} \lVert A_i - A_{i-1} \rVert$      &     Unitary     \\ 
Tether Length      &      $R_{TL}(P)  = f(s_0, s_1, ..., s_n)$      &     Additive     \\ 
Number of Contacts &      $R_{NC}(P)  = g(s_0, s_1, ..., s_n)$      &    Additive      \\ 
Azimuth            &      $R_{A}(P) = h(s_0, s_1, ..., s_n)$      &    Additive     \\ \hline
\end{tabular}
\end{table}

\subsection{Risk-Aware Reward-Maximizing Planning}
Given a viewpoint quality map as reward and motion execution risk as a function of path, the risk-aware reward-maximizing planner plans minimum-risk path to each state \cite{xiao2019explicit}, evaluates the reward collected, and then picks the one with optimal utility value, defined as the ratio between reward and risk. Executing the optimal utility path approximates the optimal visual assistance behavior. 

\section{Tethered Motion}
\label{sec::tethered_motion}
With a high-level risk-aware optimal utility path, this section presents a low level motion suite to realize the path on the tethered aerial visual assistant.

\subsection{Tether-Based Indoor Localization}
\begin{figure}
\centering
\subfloat[Localization Model]{\includegraphics[width=0.33\columnwidth]{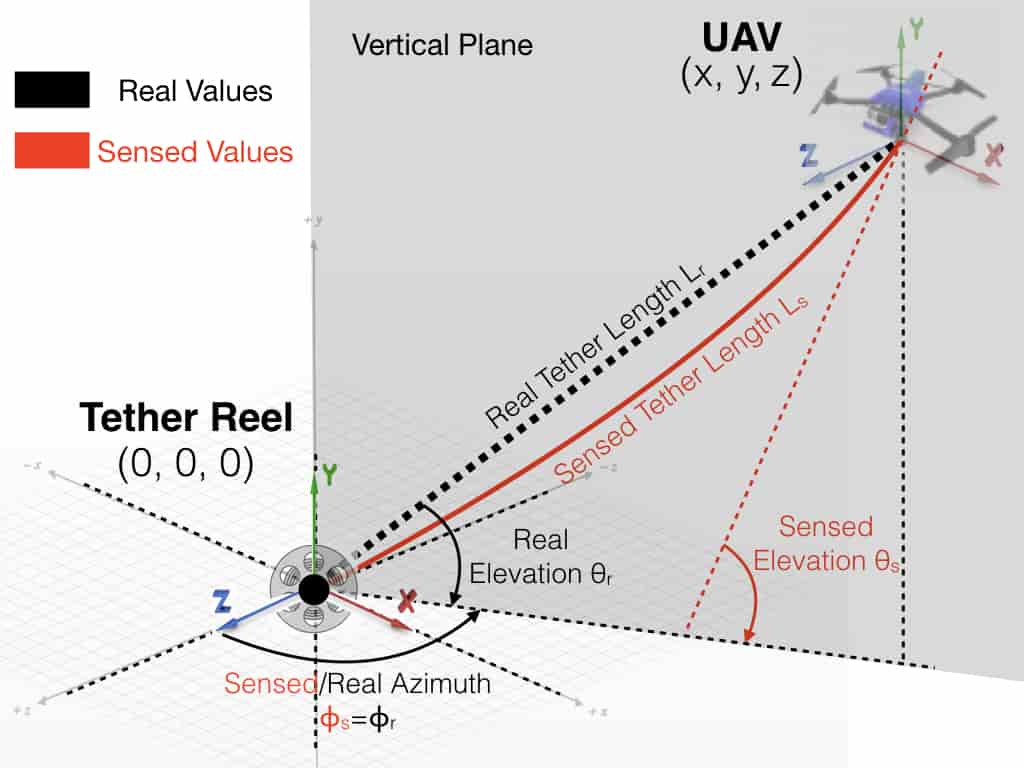}%
\label{fig::real_and_sensed}}
\subfloat[FBD of UAV]{\includegraphics[width=0.33\columnwidth]{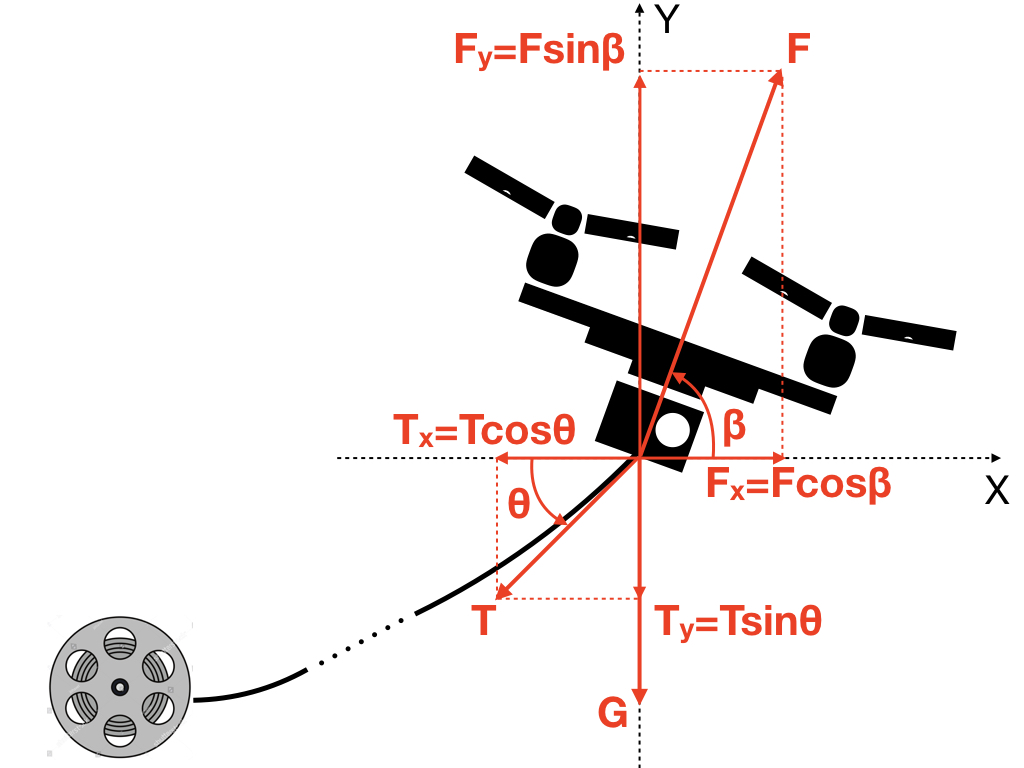}%
\label{fig::uav_fbd}}
\subfloat[FBD ofTether]{\includegraphics[width=0.33\columnwidth]{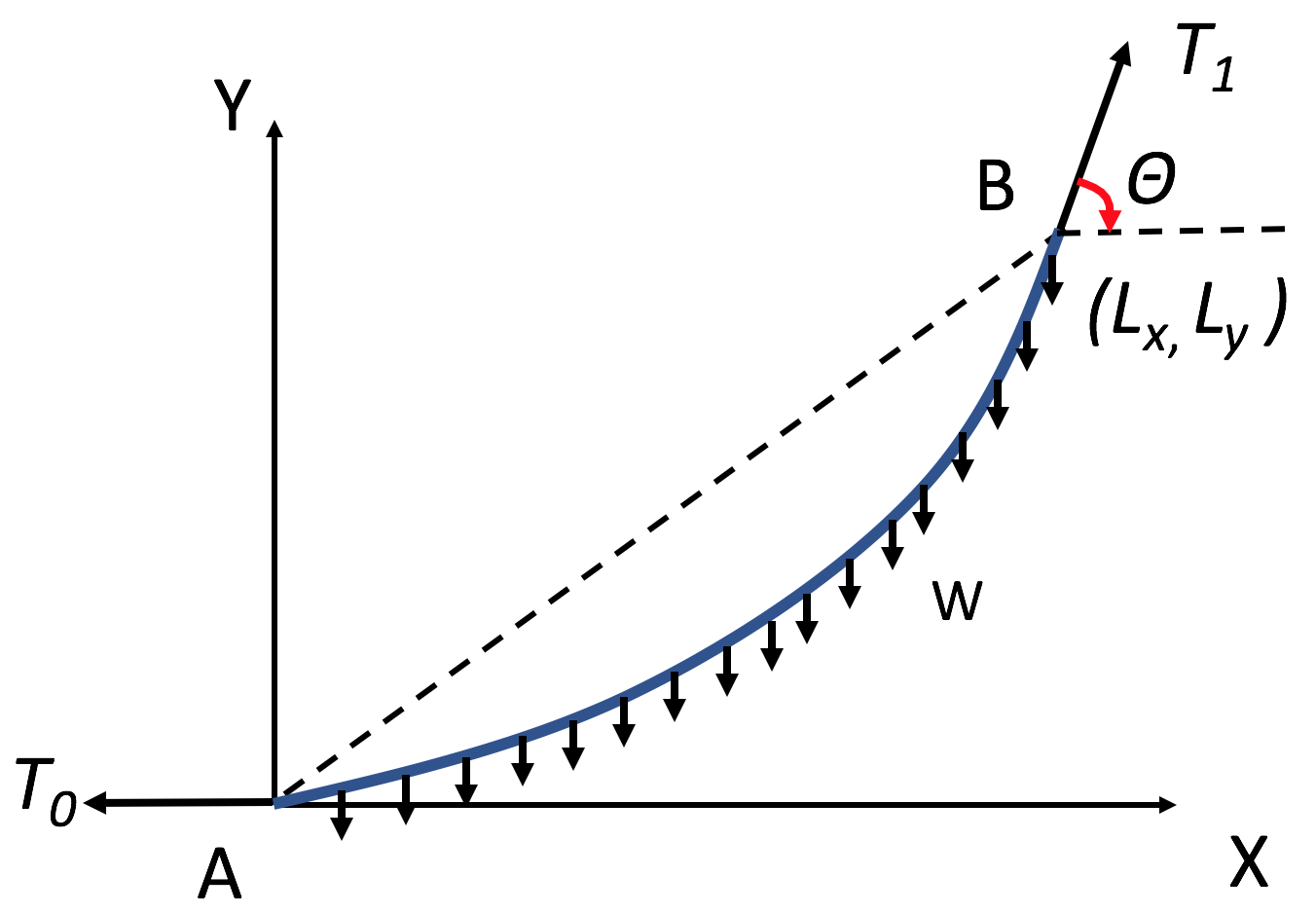}%
\label{fig::tether_fbd}}
\caption{Tether-based Localization \cite{xiao2018indoor}}
\label{fig::localization}
\end{figure}

Our aerial visual assistant uses its tether to localize in GPS-denied environments. The sensory input is the tether length $L$, elevation angle $\theta$, and azimuth angle $\phi$. The mechanics model $M$ in \cite{xiao2018indoor} corrects the preliminary localization model under taut and straight tether assumption (Fig. \ref{fig::real_and_sensed}) using the Free Body Diagram (FBD) of the UAV (Fig. \ref{fig::uav_fbd}) and tether (Fig. \ref{fig::tether_fbd}) in order to achieve accurate localization of the airframe $M: (\theta, \phi, L) \mapsto (x, y, z)$, from tether sensory input to Cartesian space location. 

\subsection{Motion Primitives}
Two types of motion primitives are used, which maps the waypoints in Cartesian space into tether-based motion commands: position control uses the inverse transformation from polar to Cartesian coordinates and three independent PID controllers to drive the position of $L$, $\theta$, and $\phi$ to their desired values (Eqn. \ref{eqn::pos_control}). On the other hand, velocity control based on the system's inverse Jacobian matrix computes velocity commands $L'$, $\theta'$, and $\phi'$ using an instantaneous velocity vector pointing from current location to next waypoint $d{\overrightarrow{\bold{x}}}/dt$ (Eqn. \ref{eqn::vel_control}). The vehicular yaw and camera pitch and roll reactively point at the center of the affordance along the entire path. 

\begin{equation}
\label{eqn::pos_control}
\left\{\begin{matrix}
L =& \sqrt{x^2+y^2+z^2}\\ 
\theta =& arcsin\frac{y}{\sqrt{x^2+y^2+z^2}}\\ 
\phi = &atan2(\frac{x}{z})
\end{matrix}\right.
\end{equation}

\begin{equation}
\label{eqn::vel_control}
\begin{pmatrix}
\frac{dx}{dt}\\ 
\frac{dy}{dt}\\ 
\frac{dz}{dt}
\end{pmatrix}
=
\begin{pmatrix}
cos\theta sin\phi & -Lsin\theta sin\phi & Lcos\theta cos\phi\\ 
sin\theta & Lcos\theta & 0\\ 
cos\theta cos\phi & -Lsin\theta cos\phi & -Lcos\theta sin\phi
\end{pmatrix}
\begin{pmatrix}
L'\\ 
\theta'\\ 
\phi'
\end{pmatrix}
\end{equation}

The vehicular yaw and camera gimbal pitch are controlled using the 3-D vehicular position localization and the 3-D Cartesian coordinates of the visual assistance point of interest. The camera gimbal roll is passively controlled to align with gravity so that the operator's viewpoint is level to the ground. Therefore, the visual assistant's camera is pointing toward the point of interest along the entire motion sequence \cite{xiao2017visual, dufek2017visual}. \cite{xiao2019benchmarking} reports detailed benchmarking results of the motion primitives. 

\subsection{Tether Contacts Planning}
\begin{figure}
\centering
\sidecaption[]
\includegraphics[width=0.6\columnwidth]{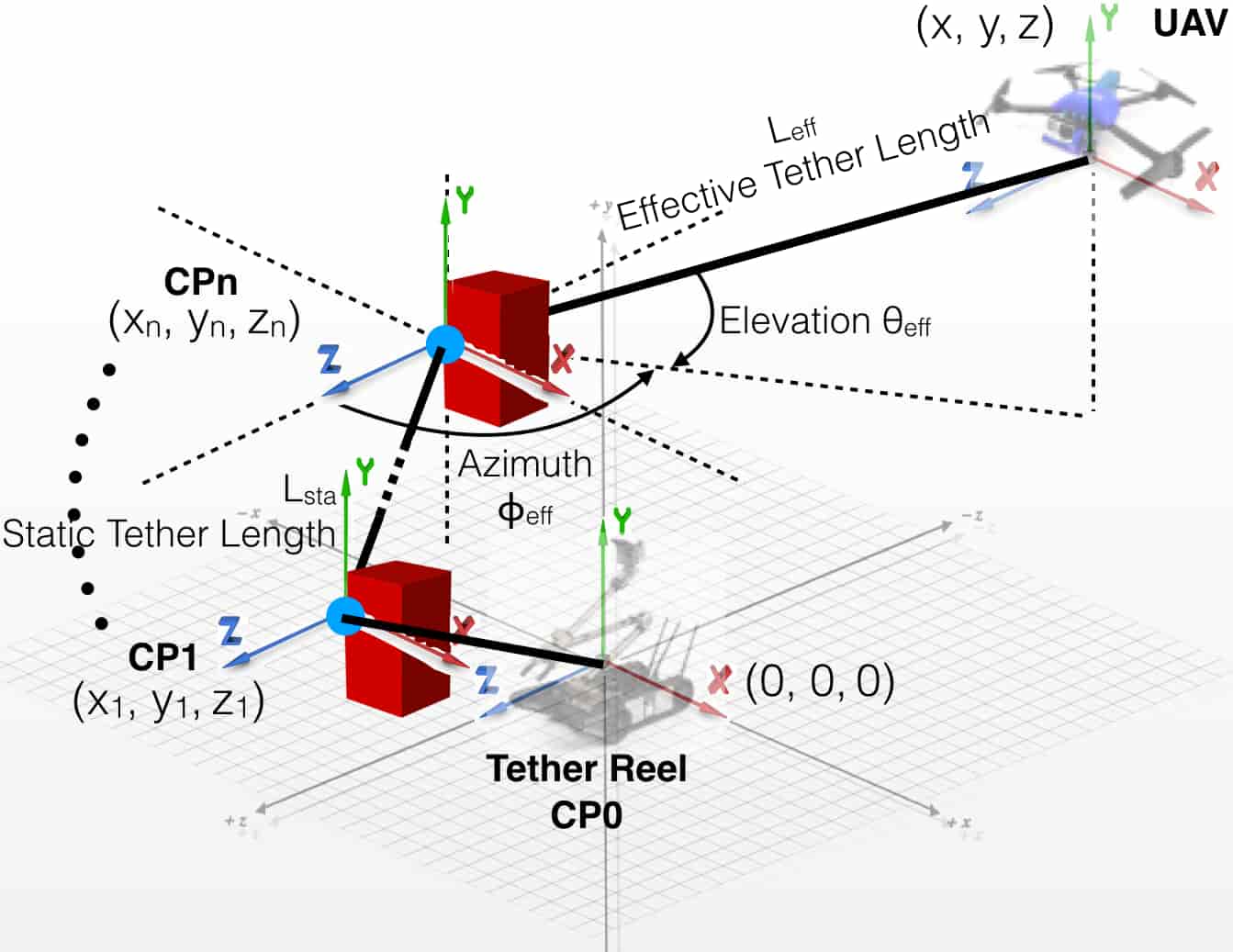}
\caption{Motion Execution with Contact(s) Planning and Relaxation: given multiple contact points along the tether, static tether length denotes the portion of the tether that wraps around the obstacles (Eqn. \ref{eqn::static_tether_length}), while the effective length is the last moving segment (Eqn. \ref{eqn::effective_controls}). Effective elevation and azimuth angles (Eqn. \ref{eqn::effective_controls}) are with respect to the last contact point ($CP_n$), instead of the tether reel ($CP_0$). }
\label{fig::contacts}
\end{figure}

In the case when some good viewpoints locate behind an obstacle and the UAV cannot reach with a straight tether, contact points of the tether with the environment are necessary. The tether contact point(s) planning and relaxation framework in \cite{xiao2018motion}, which allows the UAV to fly as if it were tetherless, is implemented on the tethered visual assistant. A new contact is planned when the current contact  is no longer within line-of-sight of the UAV, while current contact is relaxed when the last contact becomes visible again. Fig. \ref{fig::contacts} shows the motion execution with multiple contact points (CPs).  

\begin{equation}
\label{eqn::static_tether_length}
L_{sta} =  \sum_{0}^{n-1}\sqrt{(x_{i+1}-x_i)^2+(y_{i+1}-y_i)^2+(z_{i+1}-z_i)^2}
\end{equation}

\begin{equation}
\label{eqn::effective_controls}
\left\{\begin{matrix}
L_{eff} = &\sqrt{(x-x_n)^2+(y-y_n)^2+(z-z_n)^2}\\
\theta_{eff} = &arcsin(\frac{y-y_n}{\sqrt{(x-x_n)^2+(y-y_n)^2+(z-z_n)^2}})\\
\phi_{eff} = &atan2(\frac{x-x_n}{z-z_n})
\end{matrix}\right.
\end{equation}

%The final desired position control to the tethered aerial visual assistant is shown in Eqn. \ref{eqn::desired_controls}. For velocity control with contact point(s), the Jacobian matrix is computed based on the moving tether segment with respect to the last contact point. 

%\begin{equation}
%\label{eqn::desired_controls}
%\left\{\begin{matrix}
%L = &L_{eff} + L_{sta}\\
%\theta = &\theta_{eff}\\
%\phi = &\phi_{eff}
%\end{matrix}\right.
%\end{equation}

\section{System Demonstration}
\label{sec::system_demonstration}
This section presents two system demonstrations in both indoor and outdoor environments and shows the enhanced situational awareness of the operator achieved by the visual assistance. 

\subsection{Indoor Test}

\begin{figure}
\centering
\subfloat[Entering the Scene]{\includegraphics[width=0.33\columnwidth]{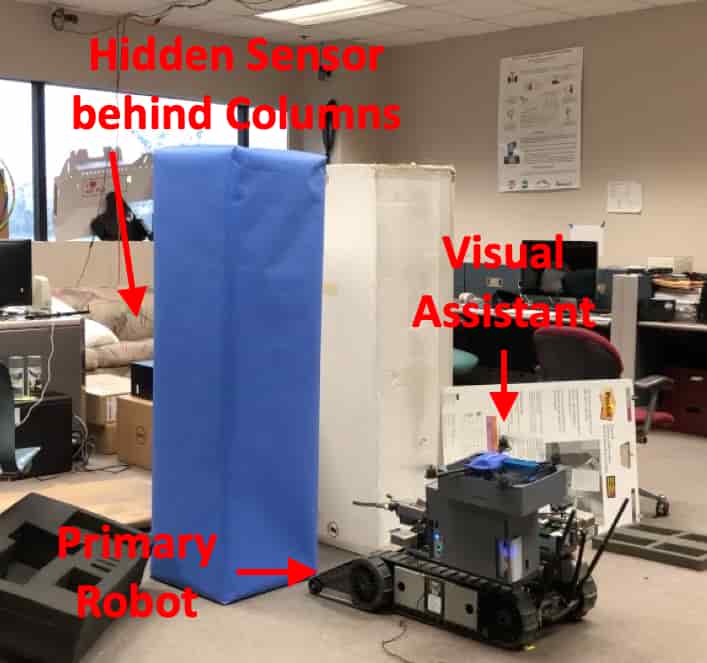}%
\label{fig::scene}}
\subfloat[Deploying for \emph{Passability}]{\includegraphics[width=0.33\columnwidth]{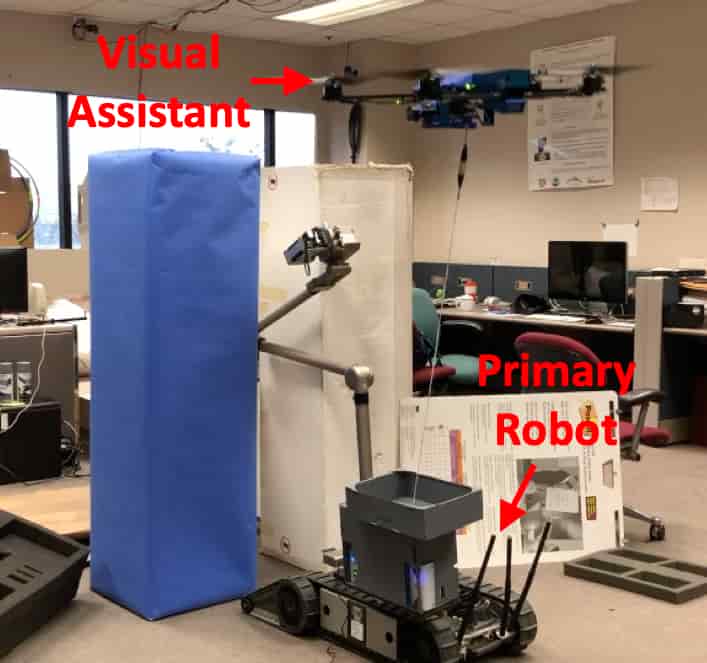}%
\label{fig::scene2}}
\subfloat[Visual Assistant View]{\includegraphics[width=0.33\columnwidth]{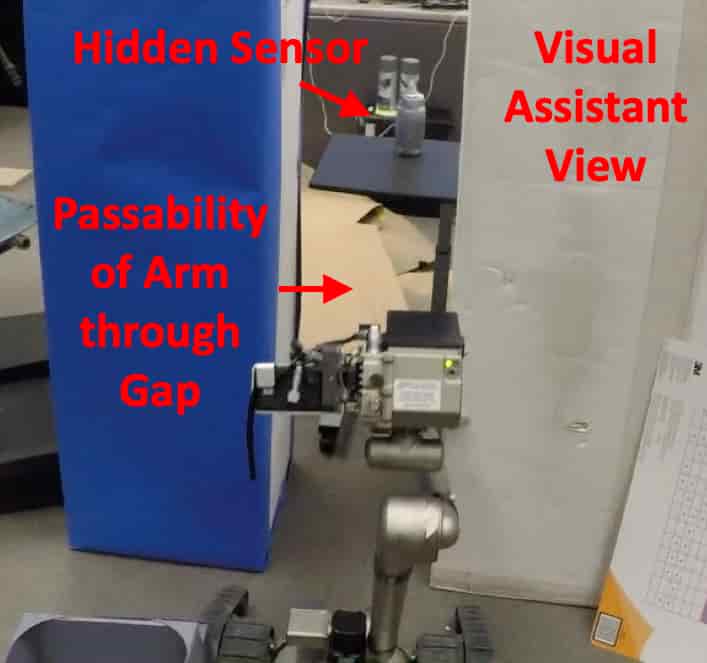}%
\label{fig::va_pass_view}}
\caption{Visual Assistance for \emph{Passability}}
\label{fig::passability}
\end{figure}

In this demonstration, the co-robots team drives into a cluttered indoor environment, with the aim of retrieving a hidden sensor. The ground robot is tele-operated and creates a map of the environment. The entry points to the sensor are all blocked by the clutter, leaving the only retrieval option through the narrow gap between the two columns (shown in blue and white in Fig. \ref{fig::scene}). Based on the viewpoint quality for \emph{passability}, the visual assistant takes off and deploys to a viewpoint from behind and above to help perceive arm \emph{passability} through gap (Fig. \ref{fig::scene2}). The visual assistant view is shown in Fig. \ref{fig::va_pass_view}, where the relative location of the arm to the narrow gap along with the hidden sensor is well perceived. 

\begin{figure}
\centering
\subfloat[Path Planning with 2 Contacts]{\includegraphics[width=0.3548\columnwidth]{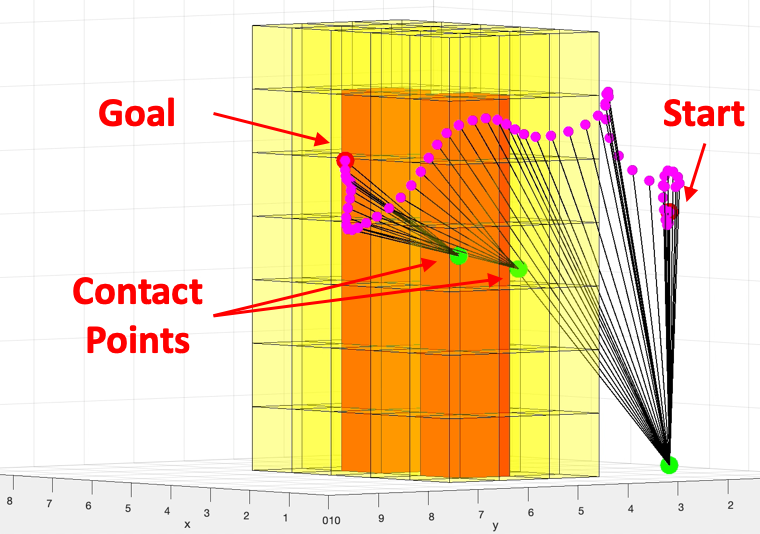}%
%0.4435
\label{fig::map}}
\hspace{10pt}
\subfloat[Deploying for \emph{Manipulability}]{\includegraphics[width=0.4452\columnwidth]{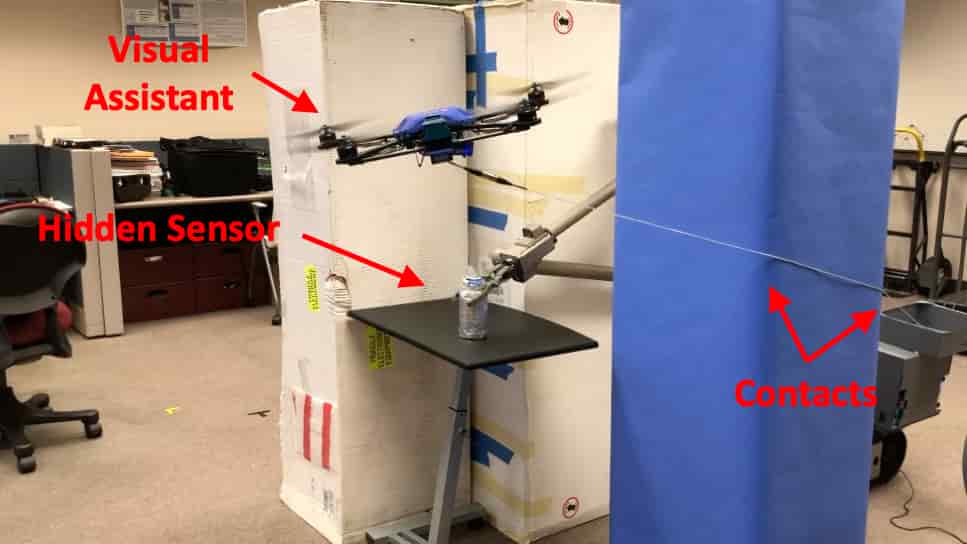}%
%0.5565
\label{fig::scene3}}\\
\subfloat[Onboard Camera and Visual Assistant View]{\includegraphics[width=0.5\columnwidth]{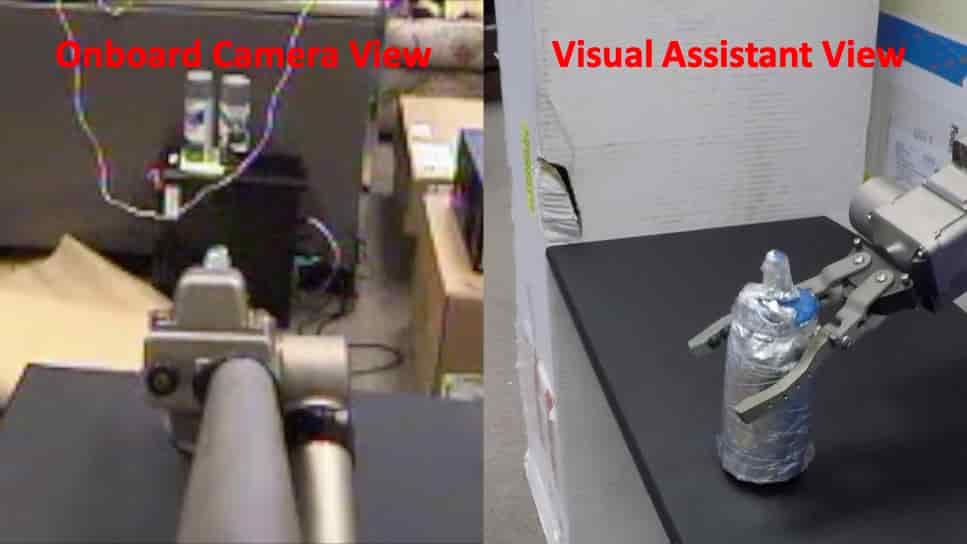}%
\label{fig::view_comparison.jpg}}
\caption{Visual Assistance for \emph{Manipulability}}
\label{fig::manipulability}
\end{figure}

After the arm passes through the gap, visual assistant switches to assist with \emph{manipulability}. Good viewpoints for \emph{manipulability} locate at the side of the gripper. After balancing the viewpoint quality reward and motion execution risk, the planner finds a goal and a path leading to it, which contains two tether contact points with the obstacles. Fig. \ref{fig::map} shows the obstacles (red), inflated space for UAV flight tolerance (yellow), waypoints on the planned path (purple), and two contact points on the obstacles (green). The tether configuration is illustrated with black lines. The actual deployment is shown in Fig. \ref{fig::scene3}. The onboard camera view on the left of Fig. \ref{fig::view_comparison.jpg} completely misses the depth perception. With this onboard view alone, the risk of not reaching or even knocking off the sensor is high. This lack of depth perception is compensated by the visual assistant view (right). 

\subsection{Outdoor Test}
\begin{figure}
\centering
\subfloat[View from Entry Point]{\includegraphics[width=0.4\columnwidth]{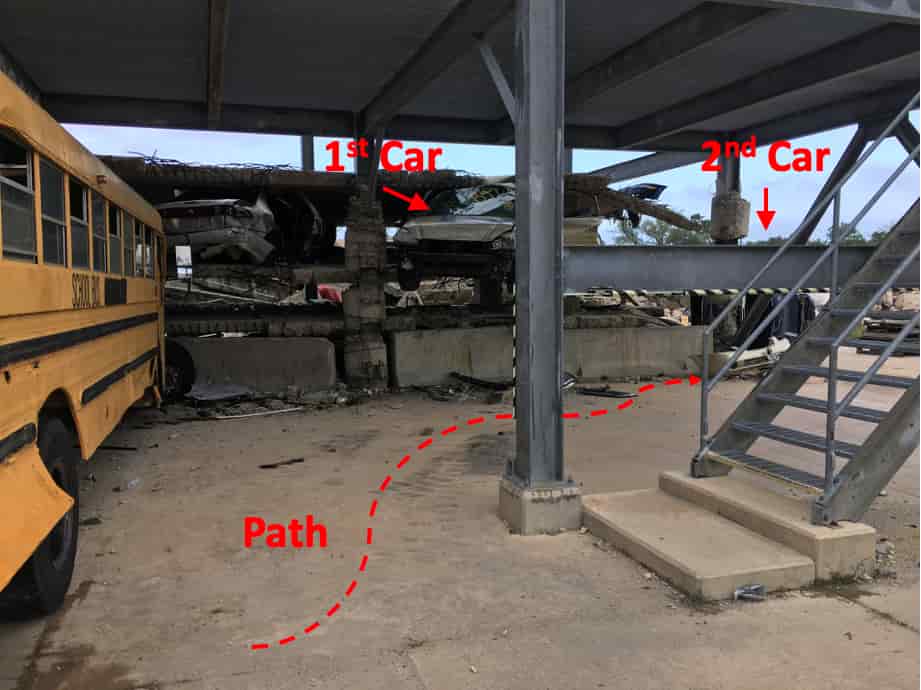}%
\label{fig::dc1}}
\hspace{10pt}
\subfloat[View from End Point]{\includegraphics[width=0.4\columnwidth]{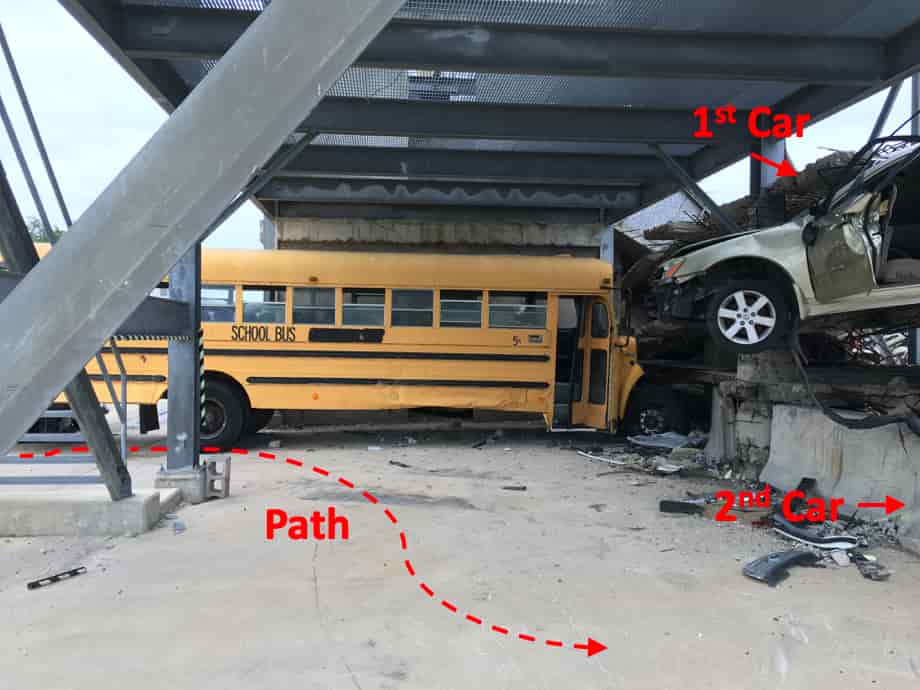}%
\label{fig::dc2}}
\caption{Disaster City \textsuperscript{\textregistered} Prop 133}
\label{fig::dc}
\end{figure}

The co-robots team is also deployed in an outdoor disaster environment, Disaster City\textsuperscript{\textregistered} Prop 133 in College Station, Texas (Fig. \ref{fig::dc}). The environment simulates a collapsed multi-story building and the mission for the co-robots team is to navigate into the building and search for victims and threats in two stranded cars. 

\begin{figure}
\centering
\subfloat[Take-off and Deployment]{\includegraphics[width=0.4\columnwidth]{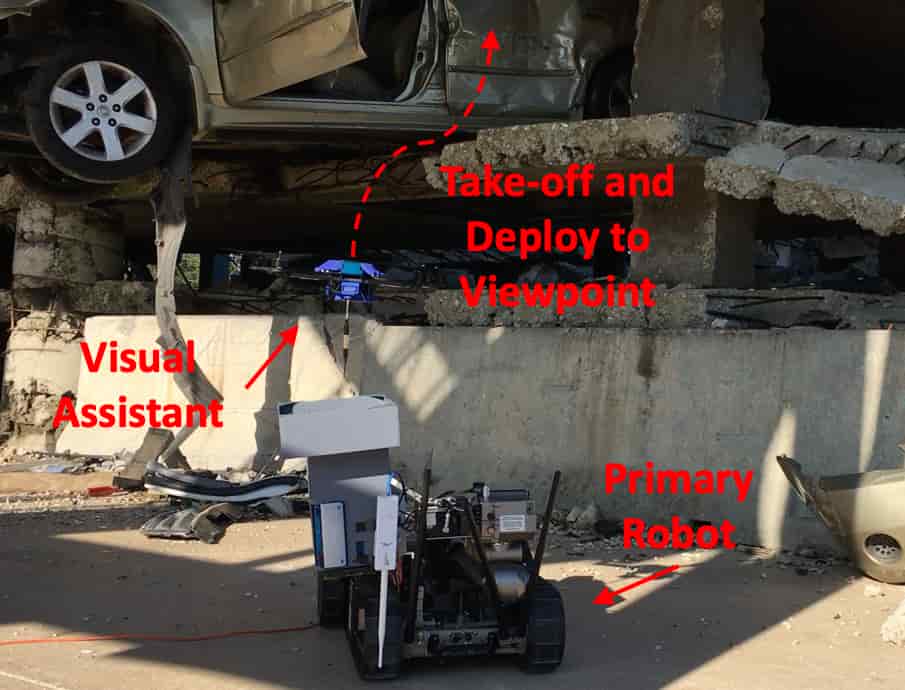}%
\label{fig::car11}}
\hspace{10pt}
\subfloat[No Victim in 1st Car]{\includegraphics[width=0.4\columnwidth]{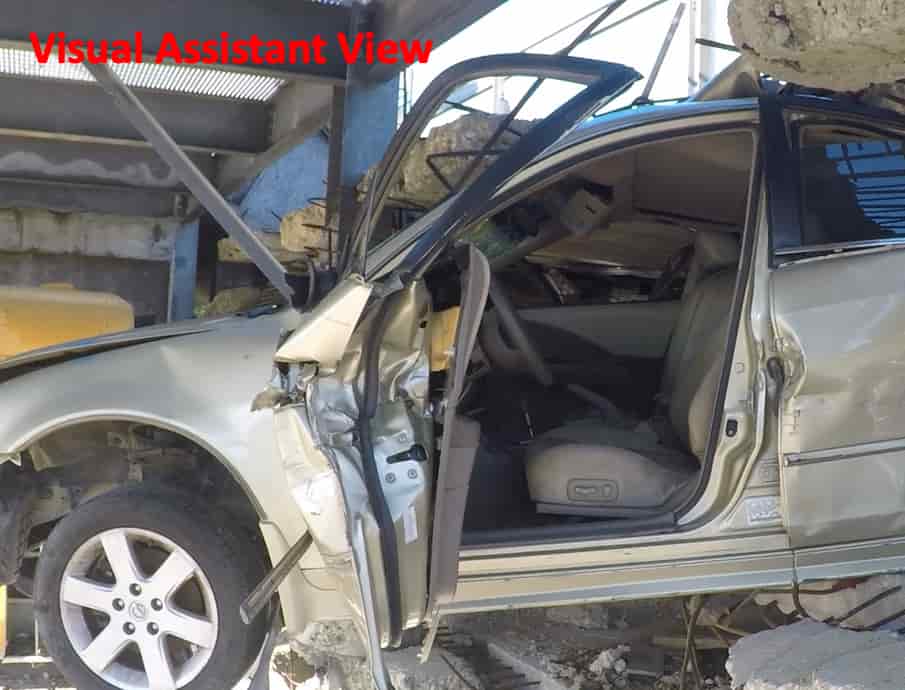}%
\label{fig::car12}}
\caption{Enhanced Coverage through Visual Assistance for 1st Car}
\label{fig::car1}
\end{figure}

After reaching first stranded car, which was on second floor but is now squeezed down by the collapse, primary robot's onboard camera is not able to reach the height to search victims inside the car. Visual assistant takes off from the landing platform and autonomously navigates to a manually specified viewpoint to look inside the car (Fig. \ref{fig::car11}). Through the elevated viewpoint provided by the visual assistant, it is confirmed that no victim is trapped in the first car. The visual assistant lands back on the primary robot and the team is tele-operated to the second car. 

\begin{figure}
\centering
\subfloat[Inspection for 2nd Car]{\includegraphics[width=0.344\columnwidth]{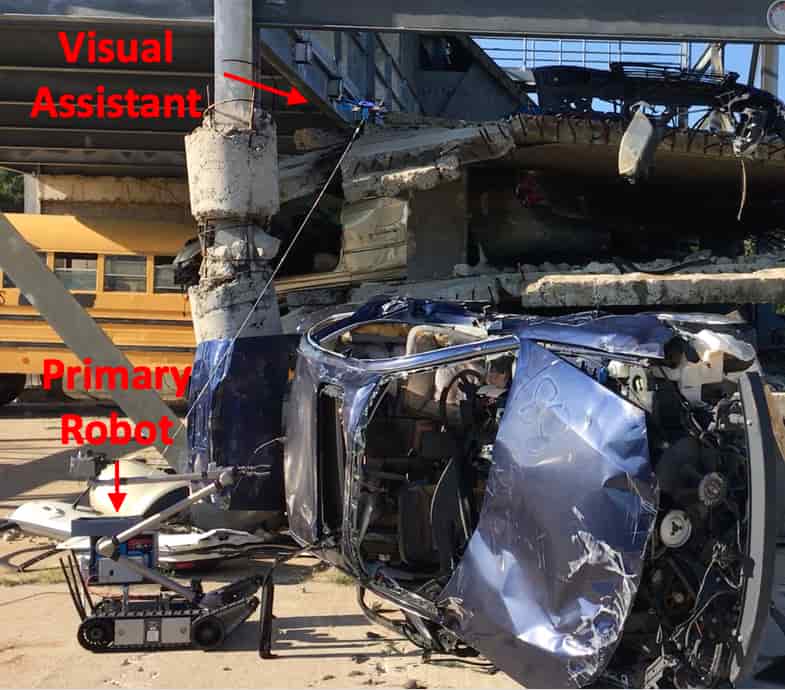}%
%0.43
\label{fig::car21}}
\hspace{10pt}
\subfloat[Assisting Insertion Depth Perception]{\includegraphics[width=0.456\columnwidth]{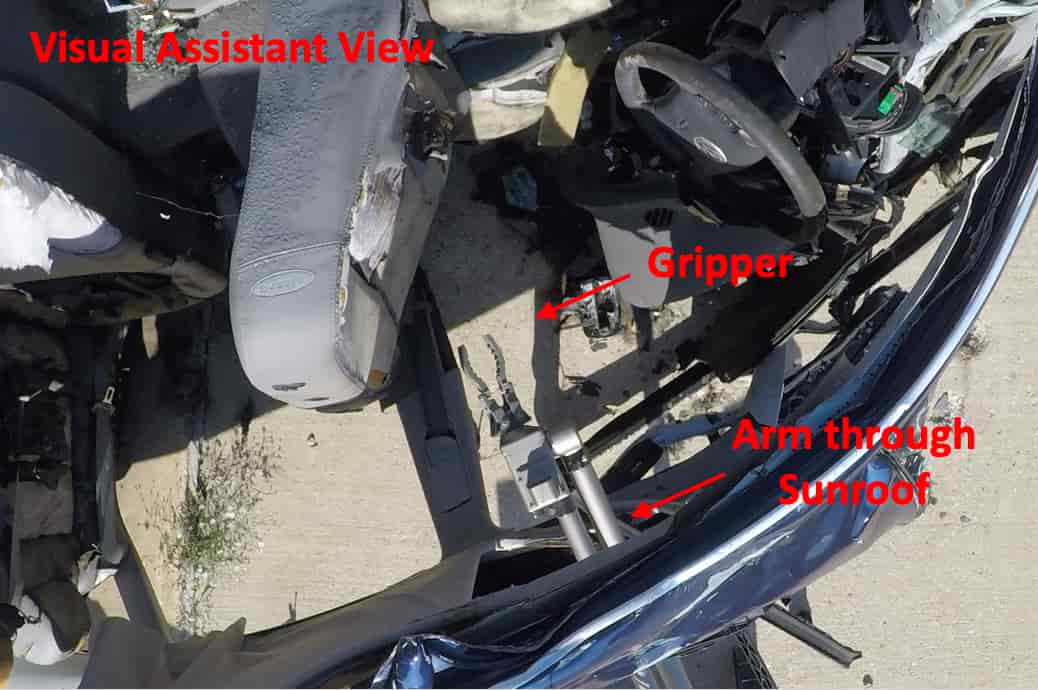}%
%0.57
\label{fig::car22}}
\caption{Car Inspection through Open Sunroof for 2nd Car}
\label{fig::insertion}
\end{figure}

The second car was tipped over during the collapse, with its sunroof open on the side. The operator intends to insert the manipulator arm into the interior for a thorough search and retrieval if necessary. For safety reasons, the goal is not automatically selected, but manually specified above the side window (Fig. \ref{fig::car21}). Looking down through the side window, the depth of the arm insertion into the car interior is clearly visible. No victim or hazardous material exist in the car. The visual assistant lands, the co-robots team finishes the mission and navigates back. 

\section{Conclusions}
\label{sec::conclusions}
We present a co-robots team equipped with autonomous visual assistance for robot tele-operations in unstructured or confined environments using a tethered UAV. The tele-operated primary ground robot projects human presence to remote environments, while the autonomous visual assistant provides enhanced situational awareness to the human operator. The autonomy is realized through a formal study on viewpoint quality, an explicit risk representation to quantify the difficulty of path execution, and a planner that balances the trade-off between viewpoint quality reward and motion execution risk. With the help of a low level motion suite, including tether-based localization, motion primitives, and contact(s) planning, the high level path is implemented on a tethered aerial visual assistant given the existence of obstacles. The co-robots team is deployed in both indoor and outdoor search and rescue scenarios, as a proof of concept of the system. Future work will focus on quantitatively measuring the performance of the co-robots team, including the reward collected, risk encountered, flight accuracy of the autonomous visual assistant, and the improvement in tele-operation of the primary robot. 

\begin{acknowledgement}
This work is supported by NSF 1637955, NRI: A Collaborative Visual Assistant for Robot Operations in Unstructured or Confined Environments. 
\end{acknowledgement}

\bibliographystyle{spmpsci}
\bibliography{references}
\end{document}